\documentclass[twocolumn,showpacs,%
  nofootinbib,aps,superscriptaddress,%
  eqsecnum,prd,notitlepage,showkeys,10pt]{revtex4-1}

\usepackage{amssymb}
\usepackage{amsmath}
\usepackage{graphicx}
\usepackage{dcolumn}
\usepackage{hyperref}
\usepackage{bm}
\usepackage{bbm}
\usepackage{float}
\usepackage{physics}
\usepackage{comment}
\usepackage{gensymb}
\usepackage{lipsum, babel}
\usepackage[dvipsnames]{xcolor}
\usepackage{cleveref}

\normalsize
\usepackage{booktabs}

\makeatletter
\def\blfootnote{\gdef\@thefnmark{}\@footnotetext}
\makeatother

\begin{document}

\title{Harvesting energy from turbulent winds with Reinforcement Learning}
\author{Lorenzo Basile$^*$}
\affiliation{The Abdus Salam International Centre for Theoretical Physics ICTP, Trieste, Italy}
\affiliation{University of Trieste, Trieste, Italy}
\author{Maria Grazia Berni$^*$}
\affiliation{University of Trieste, Trieste, Italy}
\author{Antonio Celani}
\affiliation{The Abdus Salam International Centre for Theoretical Physics ICTP, Trieste, Italy}

\blfootnote{
$^*$: Equal contribution
\\
Correspondence to: acelani@ictp.it
}

\begin{abstract}
Airborne Wind Energy (AWE) is an emerging technology designed to harness the power of high-altitude winds, offering a solution to several limitations of conventional wind turbines.
AWE is based on flying devices (usually gliders or kites) that, tethered to a ground station and driven by the wind, convert its mechanical energy into electrical energy by means of a generator.
Such systems are usually controlled by manoeuvering the kite so as to follow a predefined path prescribed by optimal control techniques, such as model-predictive control.
These methods are strongly dependent on the specific model at use and difficult to generalize, especially in unpredictable conditions such as the turbulent atmospheric boundary layer.  
Our aim is to explore the possibility of replacing these techniques with an approach based on Reinforcement Learning (RL). Unlike traditional methods, RL does not require a predefined model, making it robust to variability and uncertainty. Our experimental results in complex simulated environments demonstrate that AWE agents trained with RL can effectively extract energy from turbulent flows, relying on minimal local information about the kite orientation and speed relative to the wind.
\end{abstract}

\maketitle

\section{Introduction}
\begin{figure*}[t]
    \centering
    \includegraphics[width=0.7\textwidth]{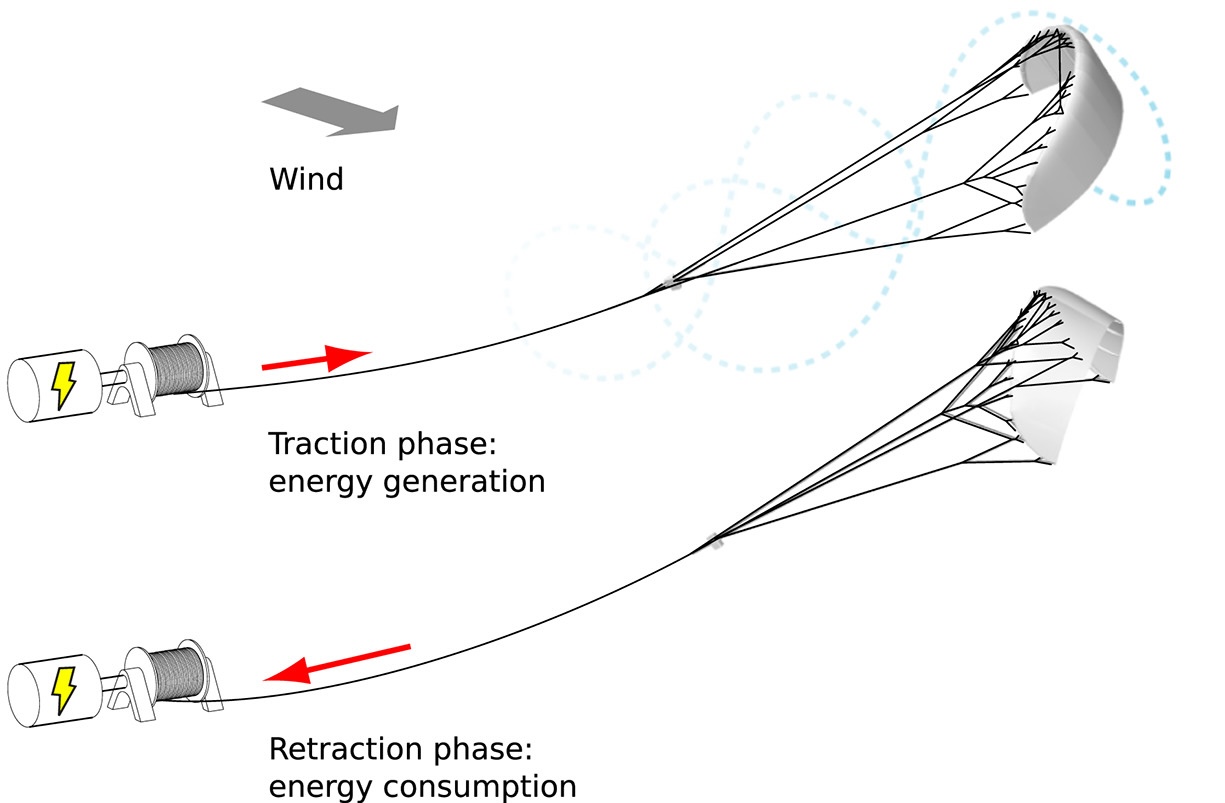}
    \caption{Schematic representation of a pumping Airborne Wind Energy system  (image adapted from Ref.~\cite{folkersma2019boundary})}
    \label{fig:schema}
\end{figure*}

In the field of renewable power sources, wind energy is particularly appealing since it can potentially power the entire world and it is largely available almost everywhere on the planet \cite{archer2005evaluation}. Currently, wind power generation happens through wind turbines, huge three-bladed devices often found in very large on-shore or off-shore wind farms. This technology has several limitations: it only allows the exploitation of low-altitude winds, known to have lower power density than high-altitude winds \cite{archer2009global}, it faces the opposition of groups who are concerned about the visual impact of wind farms on the environment \cite{diffendorfer2019geographic}, and it leads to very high costs for production and installation of the turbines \cite{elia2020wind}.

Airborne Wind Energy (AWE) is an alternative lightweight technology for wind energy harvesting, based on tethered flying devices, usually power kites. These devices fly under the effect of aerodynamic forces (drag and lift) and convert wind energy to electrical energy by means of a generator, which can be placed either directly on the device or on the ground \cite{cherubini2015airborne}.
Such technology promises to address most of the issues of traditional wind turbines, since kites can be controlled to fly at higher altitudes, where strong and constant winds can be found and since their smaller, lighter structure leads to much lower material costs and environmental impact \cite{bechtle2019airborne, archer2009global}. However, AWE technology is operationally more complex than traditional turbines and it may face reliability issues in case of critical wind anomalies \cite{salma2020improving}.

In order to safely and profitably operate an AWE system, the trajectory of the kite can be optimized by controlling its attack and bank angle through the lines that connect it to the ground station. Several works \cite{canale2009high,diehl2007optimal,gros2013control} have successfully explored the possibility of applying model-predictive control (MPC) to AWE systems. However, there are two shortcomings to this approach. First, traditional optimal control techniques are strongly model-dependent, meaning that the control strategies they provide are usually hard to generalize from one configuration to another and that the quality of the results that can be achieved depends on the quality of the model. The issue can be partially tamed by applying robust control techniques \cite{cadalen2018robust}.  Second, MPC for AWE usually focuses on minimizing the excursions from a predefined path \cite{gros2013control, fechner2018flight}, which, in general, may not be the one that optimizes power production. This lack of alignment between the objective trajectory of MPC and the final optimization objective has been highlighted by recent research \cite{song2023reaching} in the context of drone navigation, showcasing the fact that the trajectories obtained with MPC may be suboptimal in real conditions. Such problem is particularly acute in changing environments like the turbulent atmosphere, where the kite should dynamically adapt its trajectory to wind fluctuations in order to efficiently extract energy.
Complex turbulent flows are intractable for MPC methods, and one current approach \cite{crismer2024airborne} is to derive the control in mean wind conditions, and later assess its robustness in the turbulent flow.

In this work, we take a different standpoint and give a proof of concept that classical control techniques can be replaced with Reinforcement Learning (RL), which can be used even in absence of a known model, to directly aim at the maximization of energy production.

RL is a branch of Machine Learning where an agent learns to make decisions by interacting with an environment and receiving feedback in the form of rewards \cite{suttonbarto}. Unlike traditional model-based control methods, RL enables the agent to learn optimal control strategies directly by trial and error, gradually optimizing a policy that maximizes a cumulative reward signal. In the context of AWE, this reward can be designed to reflect the total energy output of the system, allowing RL to directly address the goal of maximizing power production without relying on predefined trajectories or simplified models.
RL has already proven successful in addressing flight optimization problems \cite{reddy2018glider, bellemare2020autonomous} and it has been employed for optimizing the traction phase of a pumping AWE system \cite{han2022energy} and in the context of maritime navigation powered by AWE, in which kites are used to tow a ship \cite{orzan2023optimizing}. Besides RL, Machine Learning tools have been employed for performance monitoring of AWE systems~\cite{rushdi2020power}.

Here, we focus on the optimization of whole-cycle electricity generation by means of AWE (see \Cref{fig:schema}) \cite{canale2009high}. We show that RL allows to find nontrivial control protocols which can successfully manoeuver a kite and reliably extract energy from the atmosphere, even in the strongly turbulent conditions typical of the boundary layer. We find that the energy generated in the traction phase is twice as large as the one needed to perform the retraction, on average. Remarkably, the kite operation only requires monitoring a few state variables related to its orientation and speed with respect to the local wind direction.

\section{Optimizing AWE with Reinforcement Learning}

\begin{figure*}
    \centering
    \includegraphics[width=\textwidth]{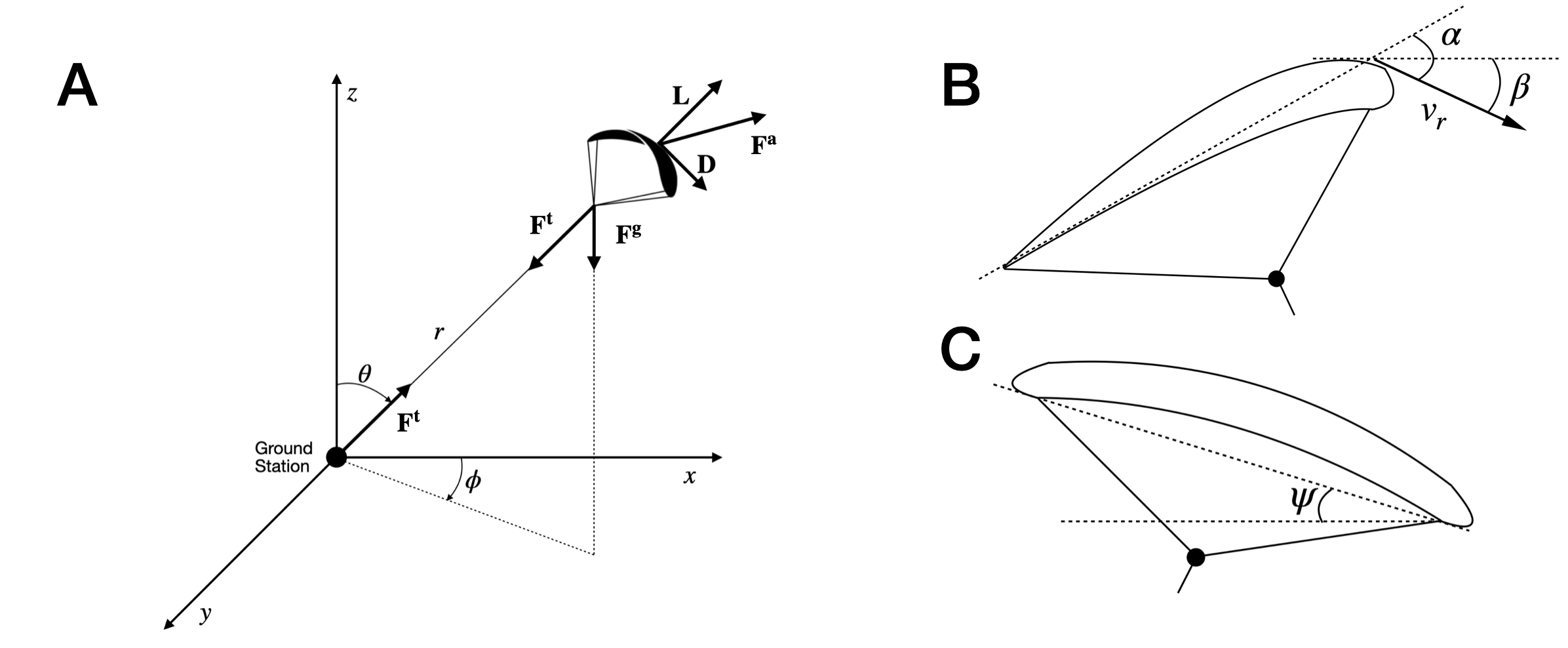}
    \caption{(A) Schematic force diagram of the AWE system. (B) Side view of the kite, showcasing the state variables $\alpha$ (attack angle) and $\beta$ (relative wind speed angle). Controlling the attack angle allows the kite to soar and glide. (C) Rear view of the kite, depicting the bank angle $\psi$, which is used to turn the kite left and right.}
    \label{diagram}
\end{figure*}
In a RL setting, a decision maker (the agent) interacts with its surroundings (the environment) and based on this interaction it receives a numerical signal which can be seen as a reward (or a penalty) for the actions it took. By trial and error, on the long run the agent understands which actions are to be preferred in each state in order to maximize the reward it obtains from the environment and optimizes its decision-making policy accordingly.

RL algorithms are traditionally categorized as either \textit{critic} methods or \textit{actor} methods. Critic methods, such as SARSA \cite{rummery1994} and Q-learning \cite{watkins1989learning}, focus on value function estimation, where the goal is to learn the expected return of states or state-action pairs (state-action value function) to indirectly guide decision-making. These methods rely on iterative updates to refine these value estimates, which implicitly define the policy by choosing actions that maximize the estimated value. In contrast, actor-only algorithms, like policy gradient methods, directly optimize the policy by adjusting its parameters in the direction that increases expected rewards. Actor-only methods do not explicitly use value functions but instead optimize the policy based on sampled returns, potentially leading to high variance in gradient estimates. While critic algorithms emphasize value function learning to inform the policy, actor-only methods focus on directly shaping the policy itself. The direct parameterization of policies in Policy Gradient methods enables them to effortlessly handle continuous action spaces.

In a combined approach, Actor-Critic methods learn approximations of both the policy and value function. This integration of both methods leverages their individual strengths for enhanced stability and efficiency in the learning process. The algorithm we adopt to control the AWE system is Twin Delayed Deep Deterministic Policy Gradient (TD3) \cite{fujimoto2018addressing}. TD3, directly derived from DDPG (Deep Deterministic Policy Gradient) \cite{lillicrap2016continuous}, is an actor-critic algorithm which employs deep neural networks to parameterize policy and value, thus falling under the umbrella of deep reinforcement learning.
Additional details regarding the setup we employ to train our RL agents can be found in \Cref{app:rl}.

In our approach, we assume that RL agents can only access three angular state variables: the attack and bank angles of the kite, and the relative wind speed angle. The attack and bank angles also serve as the controls of our system. More details on the state and control variables we consider are provided in \Cref{sec:results}.

\section{Virtual Environment}
Many configurations have been proposed for Airborne Wind Energy (AWE) systems, differing primarily in the placement of the electric machine (either on the ground or onboard the airborne device) and the type and number of airborne platforms used, such as kites, gliders, or buoyant structures. For an in-depth summary of current AWE technologies, we defer the reader to \cite{cherubini2015airborne}.

In this work, we focus on a pumping AWE system, also referred to as “yo-yo configuration” \cite{canale2009high} (\Cref{fig:schema}), in which a power kite is linked by means of a tether to an electric machine. The system operates in two primary phases: a traction (or extension) phase and a retraction (or reel-in) phase. During the traction phase, the kite moves in such a way that it unrolls the tether, putting into rotation the shaft of the electric machine, which operates as a generator to produce electrical energy. Once the tether reaches its maximum extension -- or before that, if instructed to do so -- the system transitions to the retraction phase. Here, the electric machine reverses its role to act as a motor, retracting the tether and positioning the kite for the next cycle. This cyclical motion allows the kite to repeatedly harness wind energy and generate power efficiently.
The overall goal of controlling this AWE system is to maximize energy production: 
\begin{equation}\label{energy}
\centering
E(T)= \int_t^{t+T}F(\tau)\dot{r} \, d\tau
\end{equation}
where $F$ represents the intensity of the force exerted on the tether. During the traction phase, this force amounts to the (positive) tether tension $F^t$ due to the movement of the kite, while during the retraction phase it becomes equal to the negative force applied to the tether by the motor, $F^m$.

In addition to traction and retraction phases, to improve the overall efficiency and smoothness of operation, we introduce two additional transitory phases, in accordance with previous research \cite{berra2021optimal}. These short intermediary phases facilitate a gradual shift between traction and retraction, ensuring smoother transitions and reducing mechanical stresses on the kite and tether.

Overall, our aim is to control such pumping AWE system by separately optimizing these four working phases using Reinforcement Learning.

\subsection{Kite dynamics}

Although we present a model-free approach to AWE control, it is still necessary to define a mathematical model of the system. This is because we need a simulated environment to train our algorithms.
\\Based on the model presented in \cite{canale2009high}, we represent the position of the kite using spherical coordinates  $(\theta, \phi, r)$, with $r$ being the distance of the kite from the origin (the length of the tether) and $\theta$ and $\phi$ the two angles shown in \Cref{diagram}A. Then, we can obtain a local coordinate system with unit vectors $(\mathbf{e}_\theta, \mathbf{e}_\phi, \mathbf{e}_r)$, centered in the position of the kite, whose mass we denote by $m$\footnote{Numerical values for $m$ and the other parameters of the system are provided in \Cref{app:params}.}. Applying Newton's laws of motion we obtain the following system of equations:
\begin{equation}\label{newton}
\begin{cases}
\ddot{\theta}=\frac{F_\theta}{mr}
\\
\ddot{\phi}=\frac{F_\phi}{mr\sin(\theta)}
\\
\ddot{r}=\frac{F_r}{m}
\end{cases}
\end{equation}
The total force $\mathbf{F}$ that acts on the kite is the sum of gravity force $\mathbf{F}^{g}$, centrifugal force $\mathbf{F}^{c}$, aerodynamic force $\mathbf{F}^{a}$ and line tension $\mathbf{F}^t$.
\\The key force driving the movement of the kite is $\mathbf{F}^{a}$, which is customarily decomposed in two components: lift ($\mathbf{L}$) and drag ($\mathbf{D}$). The intensities of both components (denoted by $L$ and $D$ respectively) depend quadratically on the relative wind speed $W^r$ sensed on the kite:
\begin{equation} 
\begin{cases}
L=\frac{1}{2}C_L A \rho {W^r}^2
\\
D=\frac{1}{2}C_D A \rho {W^r}^2
\end{cases}
\end{equation}
In the equation above, $A$ denotes the characteristic area of the kite and $\rho$ the air density, while $C_L$ and $C_D$ are, respectively, the lift and drag coefficients. Such coefficients are nonlinear functions of the angle of attack of the kite, one of our control inputs, and we employ their estimates provided by \cite{fagiano2009control}.
\\Drag is by definition anti-parallel to the relative wind speed, while the lift component lies in the orthogonal plane and its exact orientation is defined based on the geometry of the airfoil.
\\The full derivations for $\mathbf{L}$, $\mathbf{D}$ and all the other forces that appear in \Cref{newton} are reported in \Cref{app:model}.

\subsection{Wind pattern}\label{sec:wind}
We test our approach in a simulated turbulent Couette flow \cite{avsarkisov2014turbulent}. The flow data we employ were produced by
the authors of \cite{orzan2023optimizing} by solving the incompressible Navier-Stokes equations:
\begin{equation}
\begin{cases}
      \frac{\partial \bm{u}}{\partial t}+ (\bm{u} \cdot \nabla) \bm{u}= -\nabla p + \nu \nabla^2 \bm{u}  \\
    \nabla \cdot \bm{u}=0  
\end{cases}
    \label{eq:Navier}
\end{equation}
in a cubic domain of approximately 100 m per side. The simulation applied no-slip boundary conditions at the bottom surface, a constant streamwise velocity of $30$ m/s at the top, and periodic boundary conditions in the horizontal directions, while wind speed is not defined for $z>100$ m. This configuration, characterized by a Reynolds number of $65610$, represents a turbulent regime.

Due to the spatial limitations on the $z$ axis in the turbulent flow data at our disposal, we constrain the kite to fly at altitudes lower than $100$ m.
Besides training in the turbulent flow, we provide additional results relative to RL agents trained in a constant and uniform wind pattern, and evaluated both in the same simple setup (\Cref{app:constant}) and in the turbulent flow (\Cref{const_performance}) to assess transferability of the policy.

\section{Learning efficient strategies for traction and retraction}\label{sec:results}
An operational cycle consists of two primary phases: the traction phase and the retraction (reel-in) phase. During the traction phase, the goal is to generate the maximum amount of energy by unwinding the kite tether, allowing it to harness the kinetic energy of the wind. Once the kite tether reaches an extension of $100$ m, the kite is reeled back in, with minimal energy expenditure, and positioned appropriately for the next traction phase. 
The actual reel-in is preceded by an initial transitory phase ($T \rightarrow R$ phase) designed to reduce the radial velocity of the kite. When this radial velocity drops below a properly chosen threshold, the retraction phase can begin. 
Then, when the kite tether is almost fully rewound, if the kite is not positioned correctly, a second transitory phase ($R \rightarrow T$ phase) is carried out to position it properly.
Each phase is optimized using a separate RL agent.

We provide our agents with a small amount of information in the form of three easily measurable state variables, namely the attack angle $\alpha$ (\Cref{diagram}B), the relative wind speed angle $\beta$ (\Cref{diagram}B) and the bank angle $\psi$ (\Cref{diagram}C).
No discretization is required, as TD3 can seamlessly handle continuous state and action spaces. 
Two of the state variables (i.e., the attack and bank angles) also serve as the controls of our system, as their variation results in the possibility to control the trajectory of the kite by making it glide, soar and turn.
We constrain their values to range from $-5$° to $18$° for $\alpha$ and from $-3$° to $+3$° for $\psi$.
Any value within this range is allowed, but at each step the action taken by each agent is bounded, allowing the angles to vary only within one-degree increments or decrements. 

\begin{figure*}[!ht]
  \centering
  \includegraphics[width=\textwidth]{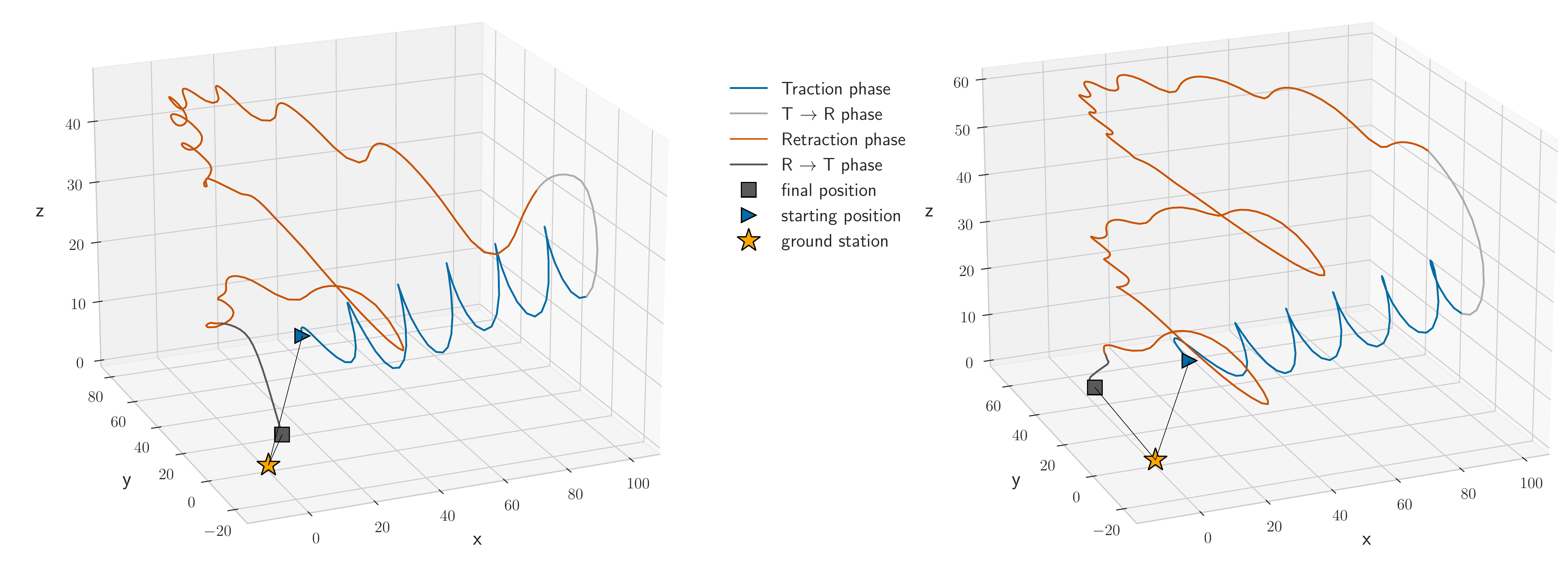}
       \caption{Sample kite flight paths in the turbulent Couette flow.}\label{fig:traj}
\end{figure*}

Sample trajectories obtained after training the TD3 agents are shown in Figure~\ref{fig:traj}

\subsection{Traction phase}
In the traction phase, the structure of the reward signal that the environment delivers to the agent is designed to maximize energy production and to keep the kite airborne: at each time step the agent receives a reward equal to the energy produced since last time step or a penalty if the kite falls to the ground (see \Cref{app:rl} for more extensive details regarding the structure of rewards and penalties for all phases).

We constrain the kite to fly at low altitudes by interrupting the traction phase when the the tether reaches an extension of $100$ m. At this point, the system is usually in a configuration that does not allow a sudden inversion of movement, as the tether is subject to high tensions and the application of an external force could lead to its breakage. For this reason, we introduce a short transitory phase to allow the retraction of the tether.

\begin{table}[!t]
  \caption{Average performance over 100 episodes in turbulent flow.}
  \label{tab:performance}
  \centering
  \begin{tabular}{lrrr}
    \toprule
    Phase     &   Duration   & Energy & Average power \\
    \midrule
    Traction & $8.08$ s & $0.057$ kWh & $25.38$ kW     \\
    $T \rightarrow R$  & $1.33$ s & $0.006$ kWh & $15.25$ kW     \\
    Retraction  & $26.01$ s & $-0.031$ kWh & $-4.35$ kW     \\
    $R \rightarrow T$  & $1.27$ s & $-0.001$ kWh & $-3.30$ kW     \\
     \midrule
    Total  & $36.68$ s & $0.031$ kWh & $2.94$ kW     \\
    \bottomrule
  \end{tabular}
\end{table}

\subsection{First transitory phase ($T \rightarrow R$)}
In this phase, the goal is to minimize the radial velocity earned by the kite during the traction phase. The emphasis is on achieving this objective without consuming energy but acting only on the attack and bank angles of the kite, while strategically positioning it in an area more suitable for the actual retraction phase.
In our simulations, this objective is achieved by moving the kite towards the zenith, akin to the ``low-power maneuver'' proposed by \cite{fagiano2009control}. This translates into reducing the $\theta$ angle compared to its value at the end of the traction phase.
The agent receives a reward proportional to a linear combination of the cosine of the $\theta$ angle and the opposite of the radial velocity of the kite. 
We assume the goal to be reached when the radial velocity of the kite is lower than a threshold $\dot{r}_{\mathrm{thr}}$. 

\subsection{Retraction phase}

Once the radial speed of the kite is lower than the threshold, the retraction phase starts. 
The generator operates as a motor, exerting a torque of magnitude $F^{m}R$, where $R$ represents the radius of the drum. 
The energy spent during each time step of the retraction phase is $\Delta E=F^{m}(t)\dot{r}(t)\Delta t$, and the overall energy is calculated summing all these contributions.

Deriving a step reward from this signal is not a suitable choice, as we would like to obtain high rewinding velocity while applying low forces.

Hence, we decide to keep the drum torque fixed, so that the goal becomes simply to rewind the tether in the shortest time possible.

To achieve this, the reward function has been designed to incentivize high rewinding velocity. Specifically, when the radial velocity of the kite is negative (indicating that the tether is being rewound), the reward is positive. This reward is structured as a linear combination of the negative radial velocity and a positive fraction derived from the effective wind speed.
Giving a high weight to the radial velocity encourages the agent to choose the trajectories with less resistance. To avoid loops, a negative reward proportional to the effective wind speed is given when the radial velocity is positive. 
A final positive reward is granted once the agent successfully achieves its ultimate goal of fully reeling-in the tether lines. This final reward is designed to have a discounted value greater than the discounted sum of intermediate rewards, and it increases with reduced energy expenditure.

The overall reward signal is summarized in \Cref{tab:reward}.

\begin{figure*}[ht!]
    \centering
    \includegraphics[width=\textwidth]{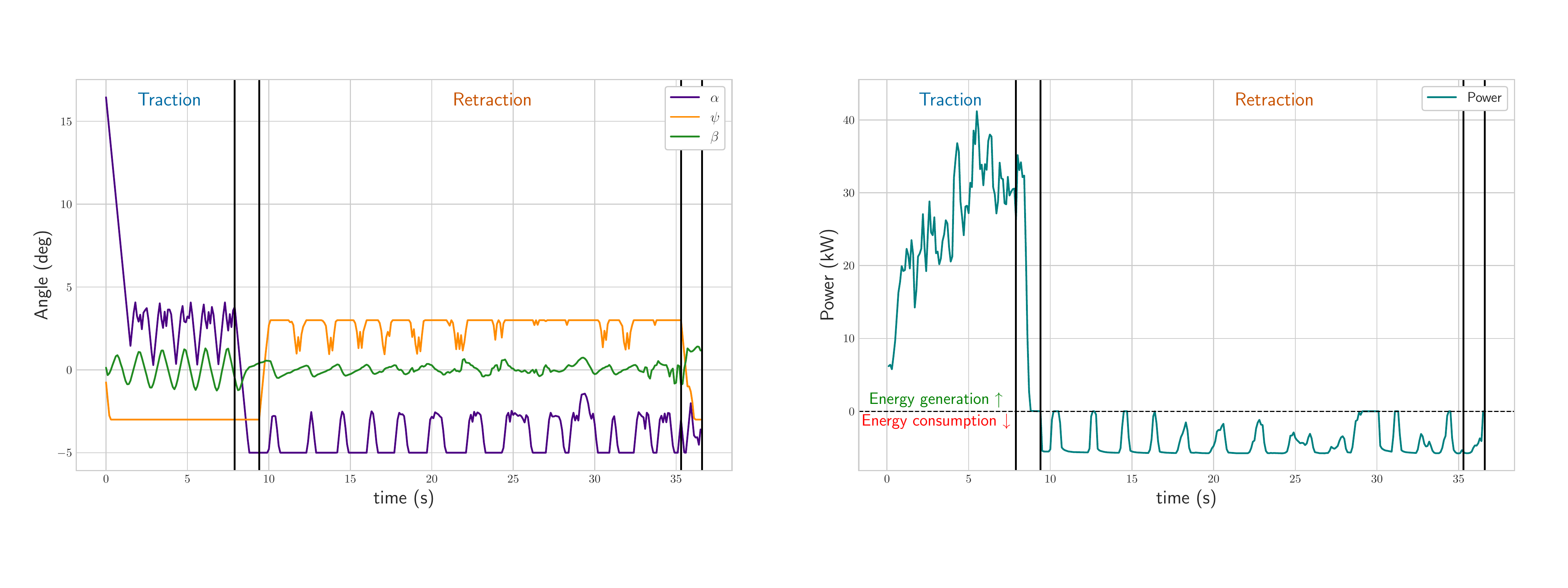}
    \caption{(\textit{left}) Control and state variables (attack angle $\alpha$, bank angle $\psi$ and relative wind speed angle $\beta$); (\textit{right}) Power profile as functions of time during an operative cycle. Both plots refer to the cycle whose trajectory is displayed in \Cref{fig:traj}-left panel.}
    \label{fig:policy}
\end{figure*}

\subsection{Second transitory phase ($R \rightarrow T$)}
The aim of this final transitory phase is to put the kite in the conditions to successfully perform a new traction phase, ensuring that the azimuthal angle $\psi$ falls within the range $[-\pi/2, \pi/2]$, and the angle $\theta$ is smaller than $\pi/4$.
During the reel-in phase, efforts were made to meet these requirements by appropriately shaping rewards and penalties. However, achieving significant results, particularly concerning the constraints on the angle  $\psi$, proved challenging and called for the introduction of this additional phase.

This phase is critical also because the kite is close to the ground. In our experiments about 44\% of realizations lead to a crash during this transitory phase, while crashes occur only in 5\% of the cases during the other phases. This stage obviously requires further work to ensure a more reliable control.

The final control agent is activated when the kite reaches a fixed threshold distance  ($r_\mathrm{thr}$) from the ground station. If the kite is already well-positioned, this control phase is skipped. The objective of the agent is to rotate the kite to bring it within the desired range by adjusting its orientation. Depending on the configuration, this may require reversing the rotation direction of the kite.

The simulation concludes once the kite achieves the desired configuration. To aid the learning process, the azimuthal angle  $\psi$  is included as a state variable, and the applied force is modulated accordingly. At each step, a reward signal is provided to encourage the rotation of the kite toward the desired orientation while continuing the rewinding of the lines. The desired configuration for the angle  $\theta$ is enforced by appropriate penalties.

\subsection{Performance of trained agents}

Even though they can access very limited state information, our TD3 agents are able to learn suitable policies to keep the kite airborne and to profitably extract energy from the wind.
During the traction phase, the agent learns to drive the kite in an approximately helical motion (as seen in the example trajectories of \Cref{fig:traj}). Given the specific wind pattern in which the kite is flying (\Cref{sec:wind}), this means that it is almost always moving crosswind. This result is in agreement with early theoretical findings that proved crosswind flight to be optimal for energy production~\cite{loyd1980crosswind}.
Overall, the retraction phase presents a more irregular trajectory and requires significantly more time than the traction phase, as reported in \Cref{tab:performance}. The table also provides quantitative results regarding energy production, averaged over 100 independent evaluation trajectories.

A realization of the learned policy over an entire operative cycle, together with the relative wind speed angle $\beta$, are reported in \Cref{fig:policy} (left panel), while the right panel reports the power produced (or consumed) by the system over the same cycle.

Our results show a significant shift in the control variables as the system transitions from the traction to the retraction phase. 
During the traction phase, the control is achieved acting only on the attack angle, which exhibits a proportional behavior with respect to the relative wind speed angle. 
The attack angle oscillates and its value is always bigger then $0$, as the goal in this phase is to produce a significant amount of lift.
The bank angle is kept constant at its lowest value. 
During the retraction phase, the attack angle is controlled to achieve a low lift, so its values fall below zero and show an approximately periodic trend (the same applies to the bank angle). The relative velocity angle in this case oscillates near zero: aligning the kite with the wind reduces both drag and lift and this makes it easier and more energy-efficient to reel-in the kite. 

\begin{figure*}[!ht]
  \centering
   \includegraphics[width=\textwidth]{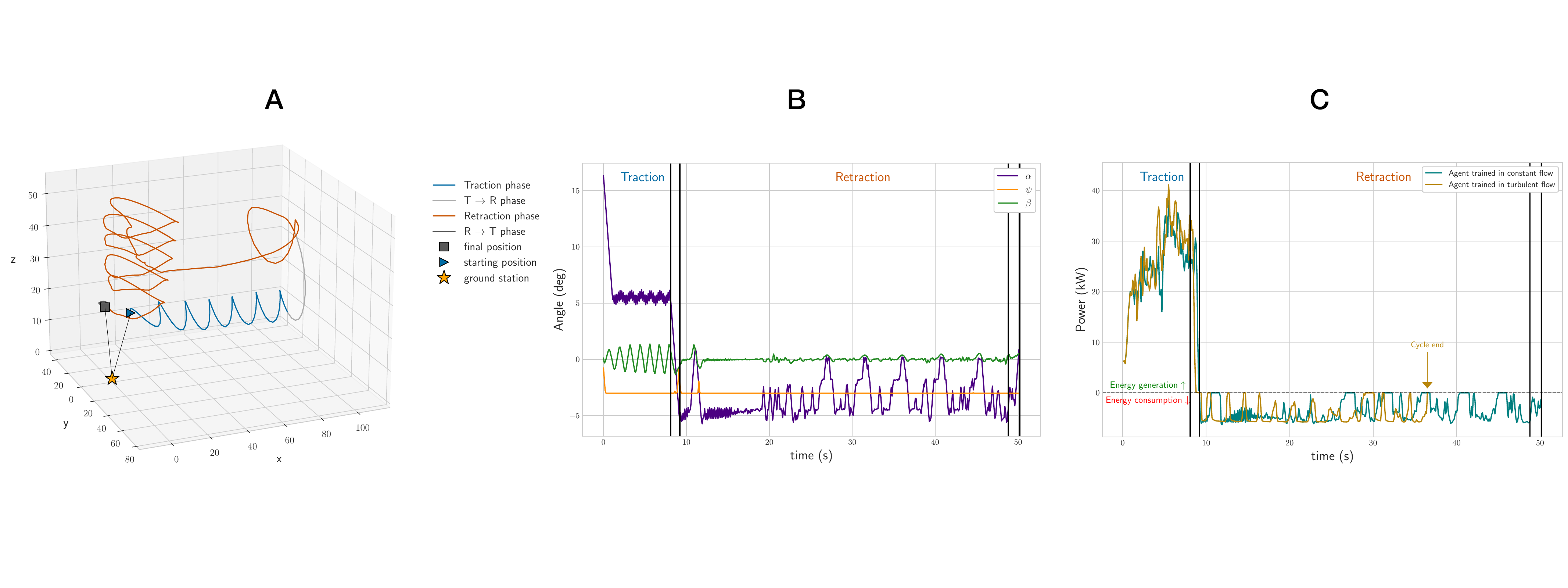}
       \caption{(A) Overall trajectory of the kite, following the policy learned in the constant wind, in the turbulent wind pattern; (B) Learned policy; (C) Power profile compared with the one obtained by an agent trained in the turbulent flow.}\label{const_turbo}
\end{figure*}

\subsection{Why turbulence matters}\label{const_performance}

Given the relative simplicity and regularity of the learned policies, one could ask whether it is really necessary to train RL agents in complex environments. To evaluate this point, we test a different approach. We train our four agents in a constant and uniform wind pattern, and assess to what extent the learned policies are successful when evaluated in the turbulent flow.

We report results for this analysis in \Cref{const_turbo}. The sample trajectory displayed in panel 5A shows that this simplified agent is still able to produce helical motion in the traction phase, while the retraction is noticeably different from the previous examples. The policy (panel 5B) mainly differs in the usage of the bank angle: in this case, it is almost always kept constant even in the retraction phase. As a consequence, in this phase the kite performs an helical descending trajectory which brings it close to the ground station. This retraction strategy is however significantly slower than the one implemented by the agent trained in the turbulent flow, making this phase longer by about $38\%$. Finally, from panel 5C we observe that the two policies are not drastically different in terms of instantaneous power production (or consumption), and the policy trained in the turbulent flow is only marginally more profitable. However, due to the extended duration of the retraction phase mentioned above, the energy gap over an entire operational cycle is significant: the constant-trained policy suffers around a $30\%$ decrease in total produced energy.

More detailed results for this experiment are reported in \Cref{app:constant}.

\section{Discussion and perspectives}\label{sec:discussion}

Our results show that RL is a viable approach to control AWE systems in a model-free way, allowing simulations in very complex environments like turbulent flow, which would be otherwise intractable with traditional optimal control methods.

An obvious limitation of our work is that we purposefully used a simplified virtual environment, from  the kite aerodynamics to the mechanics of the ground station. The wind data are also obtained from a simulation of a  turbulent Couette flow which is itself a coarse approximation of the atmospheric boundary layer. Clearly, the use of a digital twin for the virtual environment would be very valuable also in view of a simulation-to-reality deployment. 

We observed that the traction phase is easy to optimize: policies learned in turbulent and constant winds both produce rather regular helices and good performances can be achieved even with simple tabular RL algorithms (not shown here).  This is not the case for retraction
and even more so for the transitory phase at the end of the cycle ($R\to T$). 
To overcome these issues, one could design a fully automated control system that makes the transitions between phases learnable as well. This would also carry additional advantages: for instance, it is far from obvious that the end of the traction should only take place when the tether is fully unrolled.

Finally, it is worth pointing out that further work on feature selection could benefit our results. Our choice of state variables was inspired by physical considerations and by previous successful applications \cite{reddy2018glider}. However, more informative observables could be likely discovered by providing the agents with a larger feature set.

\vspace*{0.5cm}
\hrule

\appendix

\section{Mathematical model}\label{app:model}
Let us consider a cartesian coordinate system $(x,y,z)$ centered in the electric machine, with the $x$ axis aligned with the wind speed. In this system, the position of the kite can be expressed using spherical coordinates $(\theta, \phi, r)$, with $r$ being the distance from the origin (the length of the tether) and $\theta$ and $\phi$ the two angles shown in \Cref{diagram}.
\\Then, we can obtain a local coordinate system with unit vectors $(\mathbf{e}_\theta, \mathbf{e}_\phi, \mathbf{e}_r)$, centered in the position of the kite. These unit vectors can be expressed in the $(x,y,z)$ system as:
\begin{equation}
\begin{pmatrix}
\mathbf{e}_\theta
\\
\mathbf{e}_\phi
\\
\mathbf{e}_r
\end{pmatrix}
=
\begin{pmatrix}
\cos(\theta)\cos(\phi) & \cos(\theta)\sin(\phi) & -\sin(\theta)
\\
-\sin(\phi) & \cos(\phi) & 0
\\
\sin(\theta)\cos(\phi) & \sin(\theta)\sin(\phi) & \cos(\theta)
\end{pmatrix}
\end{equation}
Applying Newton's laws of motion we obtain the following equations:
\begin{equation}
\ddot{\theta}=\frac{F_\theta}{mr}
\end{equation}
\begin{equation}
\ddot{\phi}=\frac{F_\phi}{mr\sin(\theta)}
\end{equation}
\begin{equation}
\ddot{r}=\frac{F_r}{m}
\end{equation}
where $m$ is the mass of the kite.
\\
The total force $\mathbf{F}$ that acts on the kite is the sum of gravity force $\mathbf{F}^{g}$, apparent force $\mathbf{F}^{app}$, aerodynamic force $\mathbf{F}^{a}$ and line tension $\mathbf{F}^t$ (whose only nonzero component is along the $r$ axis):
\begin{equation}
\mathbf{F}=\mathbf{F}^{g}+\mathbf{F}^{app}+\mathbf{F}^{a}-\mathbf{F}^t
\end{equation}
With the simplifying assumption of a massless tether, the gravity force can be simply expressed as:
\begin{equation}
\mathbf{F}^{g}=
\begin{pmatrix}
F^{g}_\theta
\\
F^{g}_\phi
\\
F^{g}_r
\end{pmatrix}
=
\begin{pmatrix}
mg\sin(\theta)
\\
0
\\
mg\cos(\theta)
\end{pmatrix}
\end{equation}
The apparent force that comes into play is the centrifugal force, computed as:
\begin{equation}
\mathbf{F}^{app}=\mathbf{F}^{c}=
\begin{pmatrix}
F^{c}_\theta
\\
F^{c}_\phi
\\
F^{c}_r
\end{pmatrix}
=
\begin{pmatrix}
m(\dot{\phi}^2r\sin(\theta)\cos(\theta)-2\dot{r}\dot{\theta})
\\
m(-2\dot{r}\dot{\phi}\sin(\theta)-2\dot{\phi}\dot{\theta}r\cos(\theta))
\\
m(r\dot{\theta}^2+r\dot{\phi}^2\sin^2(\theta))
\end{pmatrix}
\end{equation}
\subsection{Aerodynamic force}
\label{aero}
The derivation of the equations for aerodynamic force requires some additional care. This force depends on the relative wind speed, which is computed as:
\begin{equation}
\mathbf{W}^r=\mathbf{W}^w-\mathbf{W}^k
\end{equation}
where $\mathbf{W}^w$ is the wind speed and $\mathbf{W}^k$ is the kite speed, both with respect to the ground.
\\In the local coordinate system ($\mathbf{e}_\theta, \mathbf{e}_\phi, \mathbf{e}_r$), $\mathbf{W}^k$ can be expressed as:
\begin{equation}
\mathbf{W}^{k}=
\begin{pmatrix}
\dot{\theta}r
\\
\dot{\phi}r\sin(\theta)
\\
\dot{r}
\end{pmatrix}
\end{equation}
We define now a kite wind coordinate system with basis vectors $(\mathbf{x}^w, \mathbf{y}^w, \mathbf{z}^w)$. $\mathbf{x}^w$ is aligned with $\mathbf{W}^r$ and points from the trailing edge to the leading edge of the kite, $\mathbf{z}^w$ is contained in the kite symmetry plane and points from the top surface to the bottom surface of the kite, $\mathbf{y}^w$ completes the right-handed system.
\\By its definition, $\mathbf{x}^w$ can be simply computed as:
\begin{equation}
\mathbf{x}^w=-\frac{\mathbf{W}^r}{W^r}
\end{equation}
To obtain an equation for $\mathbf{y}^w$ we first define one of our control inputs $\psi$, the bank angle of the kite, which we suppose we can control directly by acting on the length of the lines connecting the kite:
\begin{equation}
\psi=\arcsin{\frac{\Delta l}{d}}
\end{equation}
Then, if we define:
\begin{equation}
\mathbf{e}_w=\frac{\mathbf{W}^r-\mathbf{e}_r(\mathbf{e}_r\cdot \mathbf{W}^r)}{|\mathbf{W}^r-\mathbf{e}_r(\mathbf{e}_r\cdot \mathbf{W}^r)|}
\end{equation}
and
\begin{equation}
\eta=\arcsin(\frac{\mathbf{e}_r\cdot \mathbf{W}^r}{|\mathbf{W}^r-\mathbf{e}_r(\mathbf{e}_r\cdot \mathbf{W}^r)|}\tan(\psi))\label{eta}
\end{equation}
We obtain:
\begin{equation}
\mathbf{y}^w=\mathbf{e}_w(-\cos(\psi)\sin(\eta))+(\mathbf{e}_r\times \mathbf{e}_w)(\cos(\psi)\cos(\eta))+\mathbf{e}_r\sin(\psi)
\end{equation}
Finally, the unit vector $\mathbf{z}^w$ can be computed as:
\begin{equation}
\mathbf{z}^w=\mathbf{x}^w\times \mathbf{y}^w
\end{equation}
The aerodynamic force is the sum of drag and lift, that can be computed as:
\begin{equation}
\mathbf{D}=-\frac{1}{2}C_D A \rho {W^r}^2\mathbf{x}^w
\end{equation}
\begin{equation}
\mathbf{L}=-\frac{1}{2}C_L A \rho {W^r}^2\mathbf{z}^w
\end{equation}
where $\rho$ is the air density, $A$ is the kite characteristic area and $C_D$ and $C_L$ are respectively the drag and lift coefficients. These coefficients are nonlinear functions of the angle of attack of the kite, which we take as a control input, and we employ estimates provided by \cite{fagiano2009control}.
\\Finally, $\mathbf{F}^{a}$ is computed as:
\begin{equation}
\mathbf{F}^{a}=\mathbf{D}+\mathbf{L}
\end{equation}
\subsection{Tension}
In our system energy is produced by the unwinding of a tether wound on a winch, which puts into rotation the drum of an electric generator. Then, since we do not make any particular assumptions about the unwinding velocity (in \cite{canale2009high} it is controlled to be around a reference value), we have to integrate the physics of the drum into our model. To do so, we perform a simplified analysis in which we consider the interaction between drum and tether to happen on a plane orthogonal to the axis of rotation of the drum.
\begin{figure}[H]
\begin{center}
       \includegraphics[scale=0.25]{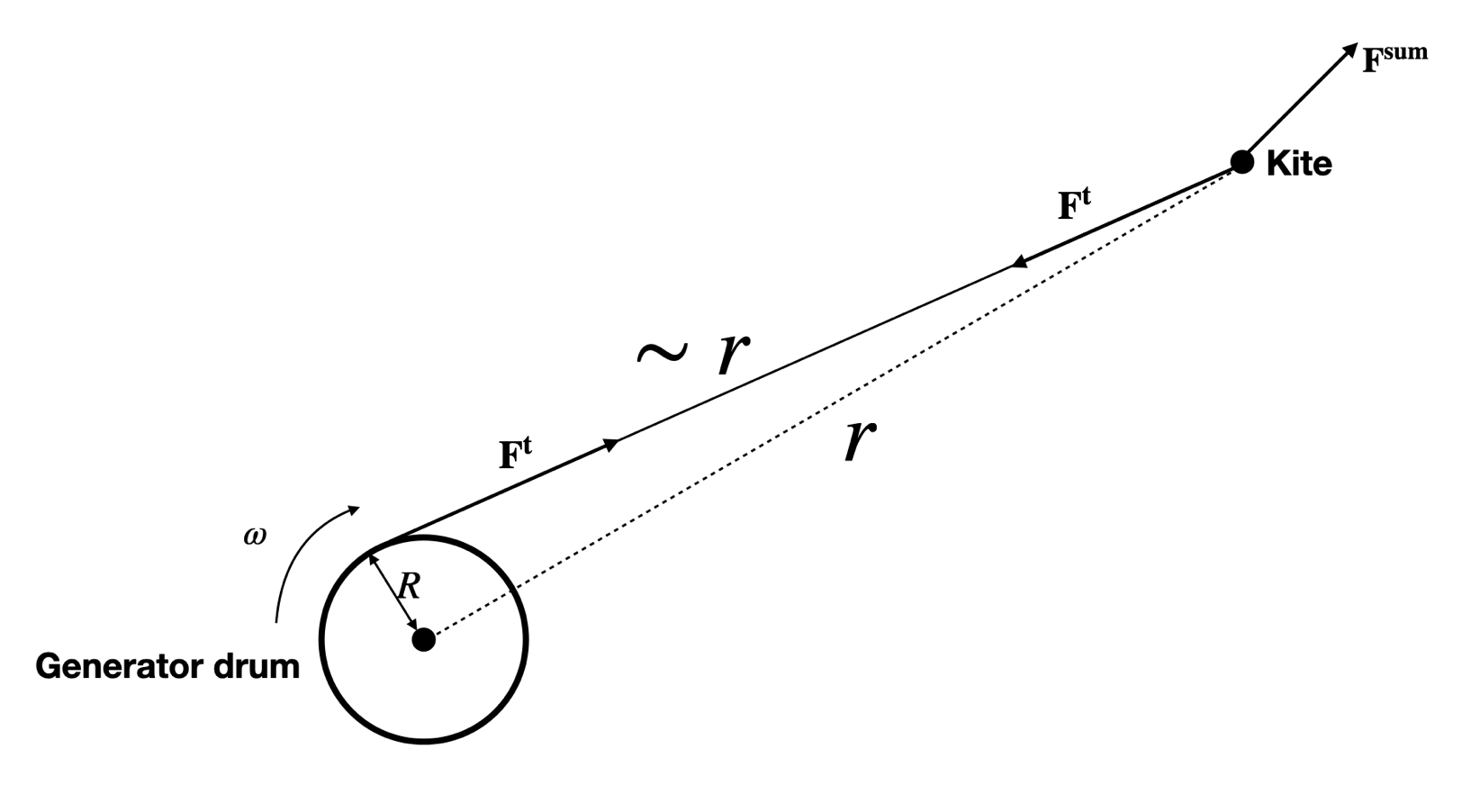}
       \caption{2D diagram for tension computation}
\end{center}
\end{figure}
\noindent When the radius $R$ of the generator drum is much smaller than the distance $r$ between generator and kite, the tension $\mathbf{F}^t$ is always directed along axis $r$ and points towards the generator. To compute its intensity $F^t$, we can apply the rotational form of Newton's laws on the shaft of the generator:
\begin{equation}
T^d-T^r=I\dot{\omega}
\end{equation}
where $\omega$ is the angular velocity of the drum, $T^d$ is the driving torque due to the tether tension, $T^r$ is the resistance torque (that we approximate as a viscous friction, proportional to $\omega$), and $I$ is the moment of inertia of the drum.
\\The angular velocity of the drum is connected to the linear expansion velocity of the tether by the equation:
\begin{equation}
\omega R=\dot{r}
\end{equation}
where $R$ is the radius of the drum.
\\Since the driving torque applied to the drum is due to the tension, Newton's equation can be rewritten as:
\begin{equation}
F^tR-k\omega=I\dot{\omega}
\end{equation}
and, substituting $\omega$:
\begin{equation}
F^tR-\frac{k}{R}\dot{r}=\frac{I}{R}\ddot{r}
\end{equation}
However, we recall that:
\begin{equation}
\ddot{r}=\frac{F_r}{m}
\end{equation}
where $F_r$ is the sum of all forces acting on the kite along the radial axis. This net force can be split into two terms: one is the tether tension $F^t$ that we are computing and the other is the net force $F^{sum}_r$ resulting from gravity forces, apparent forces and aerodynamic forces acting on axis $r$:
\begin{equation}
F_r=F^{sum}_r-F^t
\end{equation}
Then, we obtain:
\begin{equation}
F^tR-\frac{k}{R}\dot{r}=\frac{I}{R}\frac{(F^{sum}_r-F^t)}{m}
\end{equation}
The moment of inertia of a drum can be written in terms of the mass $M$ of the drum as:
\begin{equation}
I=\frac{1}{2}MR^2
\end{equation}
And, solving the equation for $T$, we obtain:
\begin{equation}
F^t=\frac{MF^{sum}_rR+2m\frac{\dot{r}}{R}k}{2mR+MR}
\end{equation}
During the rewind phase the drum will act as a generator, providing an anti-clockwise torque $C$, so that equation $(19)$ becomes: 
\begin{equation}
T^d-C-T^r=I\dot{\omega}
\end{equation}
which, following the same steps as before, leads to the tension equation: 
\begin{equation}
F^t=\frac{MF^{sum}_rR+2m\frac{\dot{r}}{R}k+2mC}{2mR+MR}
\end{equation}

\section{System parameters}\label{app:params}
In \Cref{tab:params}, we report the numerical values for the kite, engine, and algorithmic parameters that were not reported in the main text.
\begin{table}[h]
\centering
\begin{tabular}{lc}
\toprule
 Parameter & Value \\
\midrule
Kite mass $m$ & 1 kg \\
Characteristic area $A$ & 10 m$^2$\\
Air density $\rho$ & 1.2 kg/m$^3$\\
Engine drum mass $M$ & 10 kg\\
Engine drum radius $R$ & 0.2 m\\
Wind speed $W^w$ (constant flow) & 10 m/s\\
Speed threshold $\dot{r}_{\mathrm{thr}}$($T \rightarrow R$ phase) & 0.2 m/s\\
Distance threshold ${r}_{\mathrm{thr}}$ ($R \rightarrow T$ phase) & 27 m\\
\bottomrule
\end{tabular}\caption{Parameters for the AWE system under consideration.}\label{tab:params}
\end{table}

\section{RL hyperparameters and structure of rewards}\label{app:rl}

In our setup, the agent observes the state variables and selects an action every  $\Delta t = 10^{-1} \, \mathrm{s}$ , while the system evolves by integrating the dynamic differential equations described in \Cref{app:model} with an integration step of  $10^{-3} \, \mathrm{s}$ . At each decision step, a reward signal is provided to the agent to evaluate the impact of its actions on the environment. The learning process is divided into episodes, with each episode ending either upon achieving the goal or encountering a critical failure.

Each agent operates with a fixed horizon  $H$, representing the maximum number of available decision steps. While rewards guide the learning process toward the desired outcomes, penalties play a crucial role in discouraging undesirable events or behaviors.

In our setup, critical failures can occur in two scenarios. The first occurs when the kite makes contact with the ground. The second arises when the vectors  $\mathbf{e}_{r}$  and  $\mathbf{W}^{r}$  become aligned, leading to numerical instability in \Cref{eta}. In both situations, we terminate the episode and impose a failure penalty.
Additionally, once the tether reaches an extension of  $100 \, \mathrm{m}$  during the traction phase, the reel-in phase begins. However, the tether is longer than  $100 \, \mathrm{m}$ , meaning the kite is not fixed in position. If not properly managed, the tether may continue to extend even after the traction phase. To prevent this, we terminate the simulation if the tether exceeds  $130 \, \mathrm{m}$  in length or if the kite exceeds  $100 \, \mathrm{m}$  in altitude. The latter constraint arises because the available wind data is limited to altitudes below  $100 \, \mathrm{m}$.

The reward structures used for training the four agents are summarized in \Cref{tab:reward}. In the table, $E_{t_{k}}$ is the energy (expressed in kWh) gained during the $k$-th step of the traction phase, $r_{t_k}$ is the kite radial coordinate, $\dot{r}_{t_k}$ radial velocity, $W^{r}_{t_k}$ is the effective wind speed, $F^{m}$ the applied force, constant and equal to $1200$ N in our setup. 
$P(t_k)$ is a penalty scheduled according to the formula $P(t_{k}) = {-P}/{k^{0.2}}$, with $P=100$. The step reward $R_{R \rightarrow T}(t_k)$ in the second transitory phase is defined as:
\begin{equation}
    \begin{cases}
        -0.2\,\mathrm{sign}(\dot{r}_{t_k})+5(\psi_{t_{k-1}}-\psi_{t_k})\mathrm{sign}(\psi_{t_k}) & \text{if } \phi_0\psi_{t_k}\geq0\\
        -\frac{\mathbbm{1}(\dot{r}_{t_k})}{2}-\frac{\mathbbm{1}(-\dot{r}_{t_k})}{10} & \text{if } \phi_0\psi_{t_k}<0
    \end{cases}
\end{equation}
\begin{widetext}
\begin{center}
\begin{table}[h]
\centering
\begin{tabular}{lcc}
\toprule
Phase & Reward Type & Value \\
\midrule

Traction & Step Reward  & $E_{t_{k}}$ \\
         
         & Failure Penalty  & $-0.1$ \\
         & Goal Reward  & $E_{t_{k}}$ \\ \midrule
$T \rightarrow R$  & Step Reward  & $\dfrac{\cos{\theta}}{2} -\dfrac{\dot{r}_{t_k}}{10}$ \\
         
         & Failure Penalty  & $P(t_k)$\\
         & Goal Reward  & $100$ \\ 
         \midrule
Retraction  & Step Reward  & $-\dfrac{W^{r}_{t_k}}{100} \mathbbm{1}(\dot{r}_{t_k}) + \left(-\dfrac{\dot{r}_{t_k}}{10} +\dfrac{100-W^{r}_{t_k}}{200}\right)\mathbbm{1}(-\dot{r}_{t_k})$ \\
         & Failure Penalty  & $ 20-r_{t_k} $ \\
         & Goal Reward  &  $\dfrac{(HF^{m} - \sum_{i=1}^{k} F^{m})}{100}$  \\ \midrule
$R \rightarrow T$ & Step Reward  & $R_{R \rightarrow T}(t_k)$ \\
         & Failure Penalty  & $P(t_k)$ \\
         & Goal Reward  & $600$ \\ 
\bottomrule
\end{tabular}
\caption{Reward structure for all phases.}
\label{tab:reward}
\end{table}
\end{center}
\end{widetext}
The initial position of the kite is randomly sampled with the angle $\phi$ in the range [$-\pi/2$,$+\pi/2$], angle $\theta$ in the range [$\pi/12,\pi/3$] and a fixed distance of $20$ m from the ground station. The attack and bank angle are randomly initialized within their ranges of variation. 

The TD3 agents adopted in this work employ the same critic and actor network architecture as in \cite{fujimoto2018addressing}. 
A Combined Experience Replay buffer (\cite{zhang2017deeper}) is used, where the most recent experience is always sampled for performing the learning. 

All networks are trained with the Adam optimizer \cite{kingma2015adam}, using the hyperparameter choices reported in \Cref{tab:hyperparams}.

Once the training of an agent is completed, the training of the agent for the subsequent phase is conducted using the previous agent in evaluation mode (freezing the parameters of the models). This ensures that the various coordinates of the kite are automatically set, as well as the different wind parameters.

\begin{table}
\centering
\begin{tabular}{lcccc}
\toprule
 Parameter & \multicolumn{4}{c}{Phase} \\
\cmidrule(lr){2-5}
 & Traction & $T \rightarrow R$ & Retraction & $R \rightarrow T$ \\ 
\midrule
Number of episodes     & 1600    & 3000    & 90000   & 20000   \\
Critic learning rate   & 0.0001  & 0.0005  & 0.00008 & 0.00009 \\
Actor learning rate    & 0.0001  & 0.0005  & 0.00008 & 0.00009 \\
Replay memory size     & 100000  & 100000  & 100000  & 100000  \\ 
Action noise           & 0.225   & 0.25    & 0.25    & 0.25    \\ 
Target noise           & 0.2     & 0.2     & 0.2     & 0.2     \\ 
Warmup steps           & 5000    & 25000   & 25000   & 25000   \\ 
$\gamma$               & 1       & 1       & 0.99    & 0.99    \\
H                      & 3000    & 1500    & 3000    & 1500    \\
\bottomrule
\end{tabular}\caption{Hyperparameters used for TD3 \cite{fujimoto2018addressing} agents.}\label{tab:hyperparams}
\end{table}

\newpage
\section{Training in constant and uniform wind}\label{app:constant}
\begin{widetext}
    
\begin{figure}[h]
  \centering
   \includegraphics[width=\textwidth]{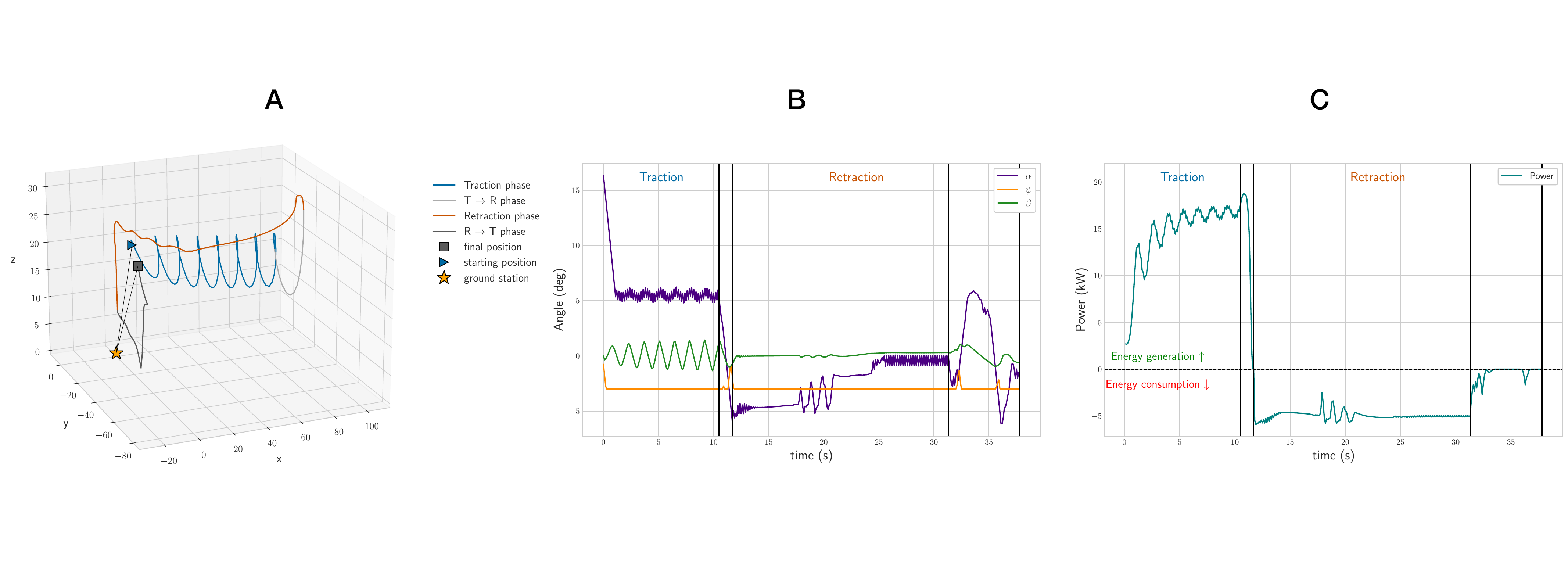}
       \caption{(A) Overall trajectory of the kite in the constant wind pattern. (B) Learned policy (C) Power profile}\label{const_wind}
\end{figure}
\end{widetext}

The same agents whose reward structure and parameters have been summarized in \Cref{app:rl}, have been trained in a constant and uniform wind pattern. In \Cref{app:performance}, we report the complete results for the policy transferability analysis introduced in \Cref{const_performance}. Moreover, in \Cref{const_wind}, we display a sample trajectory when the policy learned in the constant wind is evaluated in the same constant flow.

\begin{table}[h]
  \caption{Average performance of the agents trained in a constant and uniform wind over 100 evaluation episodes in turbulent flow.}
  \label{app:performance}
  \centering
  \begin{tabular}{lrrr}
    \toprule
    Phase     &   Duration   & Energy & Average power \\
    \midrule
    Traction & $8.25$ s & $0.055$ kWh & $23.86$ kW     \\
    $T \rightarrow R$  & $1.56$ s & $0.007$ kWh & $15.87$ kW     \\
    Retraction  & $35.91$ s & $-0.039$ kWh & $-3.92$ kW     \\
    $R \rightarrow T$  & $0.92$ s & $-0.000$ kWh & $-1.56$ kW     \\
     \midrule
    Total  & $46.65$ s & $0.022$ kWh & $1.70$ kW     \\
    \bottomrule
  \end{tabular}
\end{table}
\newpage
\bibliographystyle{abbrv}
\bibliography{main}

\end{document}